# Performance Analysis of ANFIS in short term Wind Speed Prediction

Ernesto Cortés Pérez[1], Ignacio Algredo-Badillo[2], Víctor Hugo García Rodríguez[3]

[1, 2]Department of Computer Engineering, UNISTMO University,
Tehuantepec, Oaxaca, México, C. P. 70760

[3]Department of Design Engineering, UNISTMO University,
Tehuantepec, Oaxaca, México, C. P. 70760

**Abstract**

Results are presented on the performance of *Adaptive Neuro-Fuzzy Inference system* (ANFIS) for wind velocity forecasts in the Isthmus of Tehuantepec region in the state of Oaxaca, Mexico. The data bank was provided by the meteorological station located at the University of Isthmus, Tehuantepec campus, and this data bank covers the period from 2008 to 2011. Three data models were constructed to carry out 16, 24 and 48 hours forecasts using the following variables: wind velocity, temperature, barometric pressure, and date. The performance measure for the three models is the mean standard error (MSE). In this work, performance analysis in short-term prediction is presented, because it is essential in order to define an adequate wind speed model for eolian parks, where a right planning provide economic benefits.

## 1. INTRODUCTION

The success of eolian resource forecasting depends on precision. Indeed minimizing error implies to consider factors such as: the selected forecasting model, the model parameters, the available data history, etc. Fuzzy logic and neural networks are frequently employed for estimating and forecasting [1] when available data is not sufficiently reliable because it allows tolerance levels due to the imprecision associated with linguistic terminology, which, by their very nature, are less precise than numbers. Fuzzy logic is a useful technique because it makes use of uncertainty and calibrates vagueness to find robust solutions at low computational cost [2]. Furthermore, because they are inspired by and adapted from biological systems, artificial neuronal network models simulate the cognitive processes in so much as they possess the capacity to interpret the world in the same way human beings do. Each of these techniques has its advantages and disadvantages.

A number of studies have reported successful results from the application of neural networks to wind velocity forecasts. For example, in 2009 [3] Monfared *et al.* applied a strategy based on fuzzy logic and neuronal networks to forecast wind velocity. They employed statistical properties such as standard deviation, mean, and variable calculation relation gradient as neuro-fuzzy predictor model input. They did so by using real time data obtained in northern Iran from 2002 to 2005. Readings were taken at average intervals of 30 minutes. In 2009 in average Cadenas *et al.* used one hour intervals to forecast wind velocity by using data collected by the Federal Commission of Electricity (Comision Federal de Electricidad, CFE) during a period of seven years in the La Venta, Oaxaca, Mexico [4]. This was done using a backpropagation and Madaline neuronal network with a mean square error measured at 0.0039 during the training process. In another study, Adbel Aal, *et al.* [5] performed wind velocity forecasts by using tardy neuronal networks (TTND), particularly for the maintenance of wind farms where researchers used a wind velocity data base with one-hour intervals compiled in the Dhahran region of Saudi Arabia from 1994 to 2005. The proposed model was evaluated with data from May of 2006, and the forecasting time was from 16 to 24 hours. Therefore, the network precision measurement parameter had a mean absolute error (MAE) of 0.85 m/s, and the correlation coefficient was equal to 0.83 between the actual value and the forecast. In 2009 [6] Sancho Salcedo, *et al.* employed a multilayered network with the Levenberg-Marquadt training method in which the input variables were wind velocity chosen at two points of interest with wind direction at one point, temperature at one point, and solar radiation at two points. Hence, there are six values in the input layer, a hidden layer with six neurons with a sigmoidal logarithmic activation function, and an output layer with one neuron. The forecasting time was 48 hours. The data had been collected since 2006 in Albacete province in southern Spain. The goal of this study is to perform a comparative on the performance of an Adaptive Neuro-Fuzzy Inference System (ANFIS) in forecasting time intervals of 16, 24, and 48 hours. The meteorological station at the University of the Isthmus, Tehuantepec Campus, Oaxaca, Mexico provided a data bank obtained from June 2008 to April 2011. Wind velocity readings at the meteorological station are taken every minute. However, in this research the data bank was modified by calculating averages at ten minute intervals. The following variables are considered:

- Wind velocity (*m/s*)
- Temperature (*˚C*)
- Barometric pressure (*mb*)
- Date (*hh/dd/mm*)

## 2. ARTIFICIAL NEURAL NETWORKS (ANN)

McCulloch and Pitts were the first persons to introduce a model of an elementary computing neuron. Six years later, Hebb proposed learning rules. ANN's grew rapidly, and they have been applied widely in many fields, such as pattern classification, function approximation, linear/nonlinear identification, multivariable systems. A simple ANN was composed of neuronal links that connect the neurons and assign weights and a bias to them. ANN comprises mathematical equations that mimic the brain. Since ANN is made of several neurons and different layers, it is capable of performing massive parallel computations.

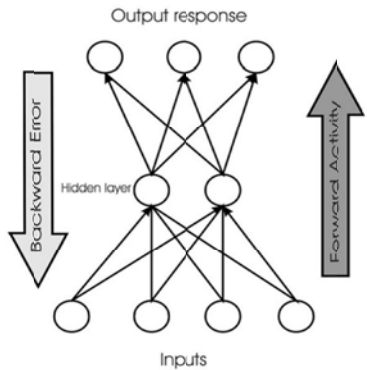

Fig. 1. Input and output of a neural network.

The position and the different neuron connections lead to ANN being classified in different groups. First feed-forward network which includes single layer perceptron, multilayer perceptron, backpropagation (BP) and radial basis functions. A second group is the recurrent networks, competitive networks, Kohonen's SOM, Hopfield network and ART models. A third group consists of the dynamic networks such as Focuses Time- Delay Neural Network (FTDNN), Distributed Time- Delay Neural Network (DTDNN), Nonlinear autoregressive network with exogenous input (NARX Network) and Layer-Recurrent Network (LRN). All of these structures have specific advantages and disadvantages. *Backpropagation* based on error backward and its correction is the most commonly used. Actually, this algorithm is based on the gradient descent method which according to error surface tries to find the best weight and bias composition in order to minimize the network error. There are two important processes in a BP algorithm: first, the error considerer calculated according to the input passed through the hidden layers of neurons; second, according to this same error, there will be backward propagation to adjust weights. However, this method has some disadvantages like slow converges, a lack of robustness, and inefficiency. One of the most successful methods which could be used to improve the training process is the Levenberg-Marquardt (LM) method which is based on both Gauss-Newton nonlinear regression and gradient steepest descent method.

## 3. FUZZY LOGIC (FL)

Fuzzy logic was thought of as a method of formalizing the kind of imprecise reasoning that humans typically perform. Everyday expressions like *it's too hot* and *it's not very high* are impossible to formulate in classical logic.

Fuzzy logic can process vague linguistic variables such as *too, very, enough* as logical formulations in computer language. Fuzzy logic fits within the framework of the multivariate logic group (there are more truth values than true and false), and it is based on fuzzy set theory.

Lofti A. Zadeh, a professor at the University of California at Berkeley founded this discipline in 1965. It came about through the application of multivariate logic to set theory. See fig. 2 for typical structure of a system based on fuzzy logic.

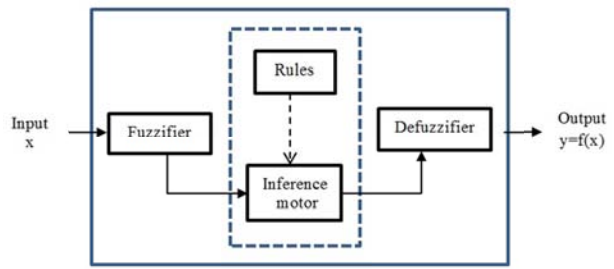

Fig. 2. System based on fuzzy logic.

## 4. ANFIS

Actually, ANFIS (*Adaptive Neuro - Fuzzy Inference System*) method is similar to a fuzzy inference system, but it uses a *backpropagation* to minimize the error. This method's performance is equivalent to both ANNs and FLs. In the case of both ANN and FL, the input passes through the input layer (by input membership function) and the output could be seen in the output layer (by output membership functions).

The parameters associated with the membership functions are modified through learning processes. The adjustment of the parameters is generated by the vector gradient. The adjusted parameters are subsequently applied to all optimization routines to reduce measurement error. Usually, if $y_t$ is the current value of period $t$ and $F_t$ is the forecast for the same period, then the error is defined as:

$$e_t = y_t - F_t \tag{1}$$

A mean square error *(MSE)* is defined as:

$$MSE = \frac{1}{n}\sum_{t=1}^{n} e_t^2 \tag{2}$$

Where *n* is the number of time periods.

ANFIS uses a combination of minimum squares and *backpropagation* for the estimation of activation function parameters. In other words, ANFIS utilizes the advantages of FL and ANN to adjust its parameters and find optimum solutions.

Both FL and ANN have their advantages. Therefore, it is good idea to combine their ability and make a strong tool which improves their weaknesses leads to minimum error. Jang [7] combined both FL and ANN to produce a powerful processing tool called ANFIS. This is a powerful tool that has both ANN and FL advantages.

Assume that the fuzzy inference system has two inputs *x* and *y* and one output *f*. For a first-order Sugeno fuzzy model, a common rule set with two fuzzy if-then rules is as follows, Jang [7]:

*Rule 1: If x is $A_1$ and y is $B_1$, then $f_1 = p_1 x + q_1 y + r_1$*

*Rule 2: If x is $A_2$ and y is $B_2$, then $f_2 = p_2 x + q_2 y + r_2$*

Let the membership functions of fuzzy sets $A_i$, $B_i$, $i = 1, 2$, be $\mu A_i$, $\mu B_i$. In evaluating the rules, choose *product* for T-norm (logical *and*).

1. Evaluating the rule premises results in:

$$w_i = \mu A_i(x)\, \mu B_i(y),\ i=1,2. \tag{3}$$

2. Evaluating the implication and the rule consequences gives:

$$f(x, y) = \frac{w_1(x, y)f_1(x, y) + w_2(x, y)f_2(x, y)}{w_1(x, y) + w_2(x, y)} \tag{4}$$

Or leaving the arguments out:

$$f(x, y) = \frac{w_1 f_1 + w_2 f_2}{w_1 + w_2} \tag{5}$$

This can be separated into phases first by defining:

$$\overline{w_i} = \frac{w_i}{w_1 + w_2} \tag{6}$$

Then *f* can be written as:

$$f = \overline{w_1} f_1 + \overline{w_2} f_2 \tag{7}$$

All computations can be seen in fig. 3 and 4.

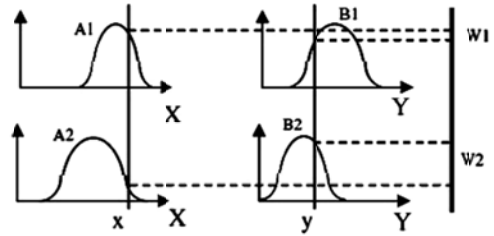

Fig. 3. First-order Takagi-Sugeno Fuzzy Model.

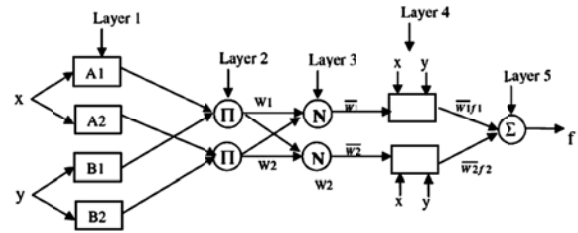

Fig. 4. Equivalent ANFIS architecture with two inputs and one output.

## 5. DATA COLLECTION

The data for this study was provided by a meteorological station located at the Tehuantepec campus of the University of the Isthmus. This data base covers the period from June 2008 to April 2011. It contains information on a number of variables such as wind velocity, temperature, solar radiation, barometric pressure, humidity, wind velocity, direction, etc. It is worth mentioning that, although wind velocity is the variable being analyzed in this research, more information may be encountered in another time series. This data may be used for more exact forecasts by using what is called variable intervention or indicator intervention, which represents additional data in the time series or information about the period in which the forecasting is realized.

It is very useful to review the values of other variables when creating a forecasting model, for example the relationship between the variable to be predicted and other variables. An information additional or other variable is usually selected to fall within the same time interval relative to the variable to be predicted.

Some of the additional variables (along with wind velocity in *m/s*) to be considered are:

- Temperature (˚*C*)
- Barometric pressure (*mb*)
- Date (*hh/dd/mm*)

Nevertheless, more information does not always mean better forecasts. Sometimes this can degrade ANFIS characteristics; such as teaching, learning, generalization, and forecasting. It is always necessary to generate relevant information for the ANFIS, provided that this is possible. Wind velocity in the Isthmus of Tehuantepec is shown in figure 5.

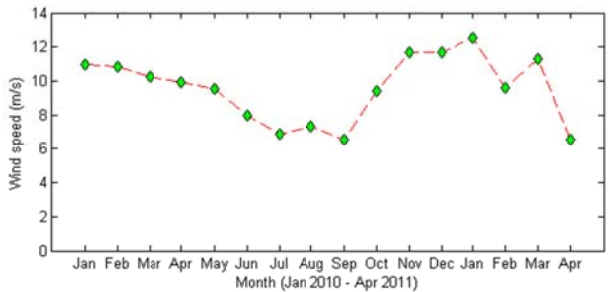

Fig. 5. Wind speed from January 2010 to April 2011 in the Isthmus of Tehuantepec.

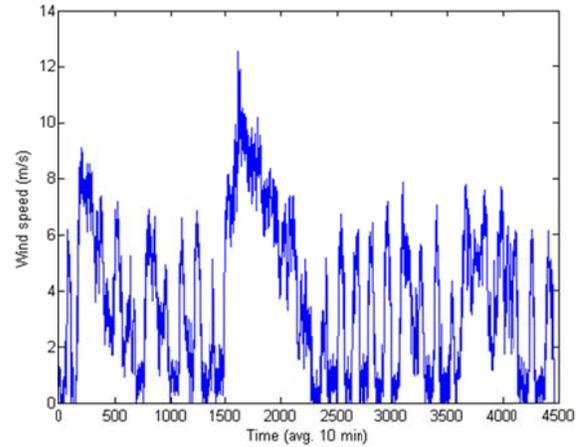

Fig. 6. Ten-minute interval wind speed averages from January 2011 in the Isthmus of Tehuantepec.

## TIME SERIES: MODELED FOR NEURO-ADAPTIVE FUZZY INFERENCE SYSTEM (ANFIS)

The time series forecast using an ANFIS consists of teaching the network the history of a variable in a selected time interval, thereby indicating information to be learned in the future. Past data is entered as input into the network and data represent future ANFIS output. An ANFIS is capable of predicting different kinds of data; nevertheless, the focus of this investigation is to forecast a time series of wind velocity in the Isthmus of Tehuantepec.

Wind velocity series demonstrate the development of a value over time. Other factors such as barometric pressure, humidity, temperature, and solar radiation, can influence this value. With the time series it is necessary to be familiar with the values at point *t* in order to make forecasts at point (*t* + *P*). Hence, it is also necessary to create a map with *D* sample of data points every Δ time unit. Thus, the following notation is used:

$$y(t+p) = [x(t_{D-1} - \Delta), ..., x(t_1 - \Delta), x(t)] \quad (8)$$

Sample for estimating 100 forward periods with four sample values: *D = 4, Δ = 1, P= 100*.

$$y(t+100) = [x(t_3 - 3), x(t_2 - 2), x(t_1 - 1), x(t)] \quad (9)$$

This method is iterative. The construction of a matrix with *W* samples is desired and the number of examples depends on the duration of the time series.

The wind speed readings were taken every minute at the meteorological station from June 2008 to April 2011.

## 6. TRAINING OF ADAPTIVE NEURO-FUZZY INFERENCE SYSTEM (ANFIS).

There are several training steps involved in ANFIS. During the first step, the form of initial fuzzy sets is determined. The number of fuzzy sets and the universe of discourse depend directly on the number and range of the variables. That is to say, a large variable set requires a large number of fuzzy sets as input. In general, there is no method for determining the ANFIS parameter values. However, in accordance with a number of tests performed and the author's recommendation [3, 4, 5, 6, 7, 8, 9, 10], the form of the fuzzy sets is as shown in figure 8, and the number of input sets was established as follows:

- Temperature: 81 Gaussian sets
- Barometric pressure: 81 Gaussian sets
- Date: 81 Gaussian sets
- Wind velocity: 81 Gaussian sets

The Gaussian and Bell sets were created in accordance with the following functions:

$$gaussian(x; \sigma, c) = e^{\frac{-(x-c)^2}{2\sigma^2}} \qquad (10)$$

$$bell(x; a, b, c) = \frac{1}{1 + \left|\frac{x-c}{a}\right|^{2b}} \qquad (11)$$

Forming a total of 81 rules with the following structure:
1. If $d_1$ is $D_1$ and $p_1$ is $P_1$ and $t_1$ is $T_1$ and $w_1$ is $W_1$, then $y$ is $Y_1$

2. If $d_2$ is $D_2$ and $p_2$ is $P_2$ and $t_2$ is $T_2$ and $w_2$ is $W_2$, then $y$ is $Y_2$

3. If $d_3$ is $D_3$ and $p_3$ is $P_3$ and $t_3$ is $T_3$ and $w_3$ is $W_3$, then $y$ is $Y_3$

4 ...

81. If $d_{81}$ is $D_{81}$ and $p_{81}$ is $P_{81}$ and $t_{81}$ is $T_{81}$ and $w_{81}$ is $W_{81}$, then $y$ is $Y_{81}$

Where:

$D_i$: date, $P_i$: barometric pressure, $T_i$: temperature, $W_i$: wind velocity, y $Y_i$: forecast (wind velocity)

The ANFIS structure is as follows:

- Number of input values: 4 (temperature, barometric pressure, date, wind velocity)
- Number of output values: 1 (wind velocity)

It is necessary to mention that the number of input sets and the number of rules to be constructed increase if the number of variables used to perform the forecast increases.

The foregoing structure can be observed in figure 7.

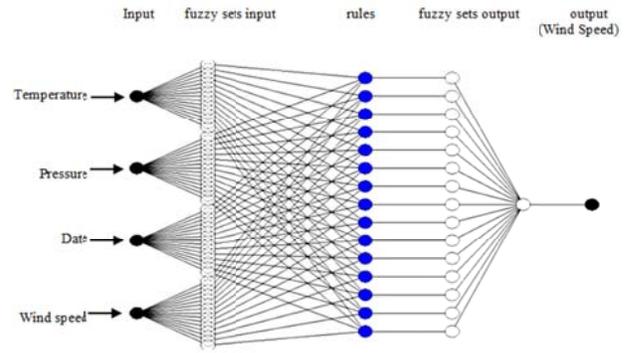

Fig. 7. ANFIS structure.

To carry out training, three data models were constructed to be incorporated into the ANFIS. Minimization of error was attempted by means of the learning process. The three data models were constructed to make forecasts within 16 hours, 24 hours, and 48 hour time forward periods.

It is important to monitor and verify input as it goes through the network learning process while keeping process parameters in mind and making necessary adjustments to minimize error. Sometimes a small error is enough to cause the network to perform poorly which can result in overtraining. Another way to arrive at a solution is to establish stoppage criteria in the training phase when a maximum acceptable error is typically established. This is done to maximize the network's generalization.

Six epochs were tested with the three data models to obtain a decrease in the mean standard error *(MSE)* and to achieve stability during training with a tolerance level of 0.00001. The *Takagi-Sugeno* fuzzy inference system was trained to estimate the results produced by the three models.

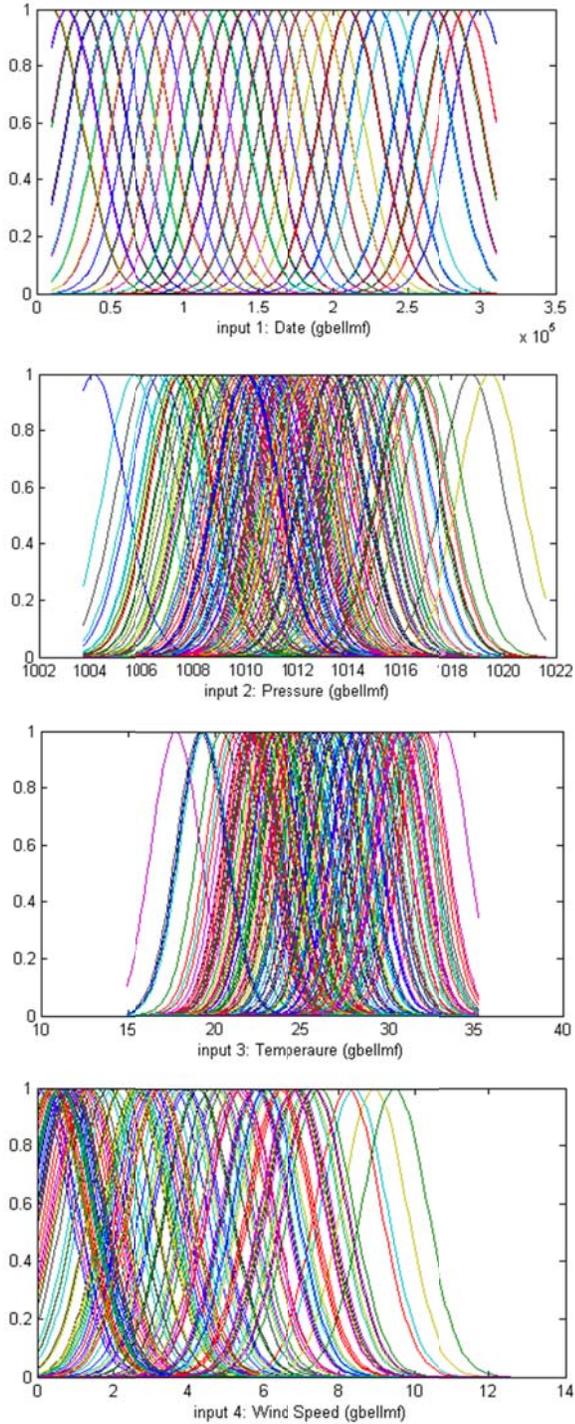

Fig. 8. Input variables and their shapes.

The results obtained for the 16 hours forward forecast are shown in figure 9. A lineal regression, with a result of *r*=81.1185, was applied in order to discover how close to the original data the data obtained during the forecast were.

Figure 10 shows the ANFIS forecast for a 24 hours forward forecast with *r*=80.8011. This value is less than the one produced for the 16 hours period, in which a partial inference can be made suggesting, that the error increases as the forecast intervals are longer.

The ANFIS forecast for a 48 hours forward forecast with *r*=77.0955 can be observed in figure 11. This value is less than the ones from the 16 hours and 24 hours forecasts, in which it is shown that the error increases with longer forecasts periods. Table 1 shows the results of the tests.

Table 1: Performance of ANFIS for 16, 24 and 48 hours periods.

| Time | r% | Epoch | Training | MSE |
|---|---|---|---|---|
| 16 h | 81.118 | 6 | 23 min. | 1.633 |
| 24 h | 80.801 | 6 | 2.7 min. | 1.569 |
| 48 h | 77.095 | 6 | 6.01 min. | 1.568 |

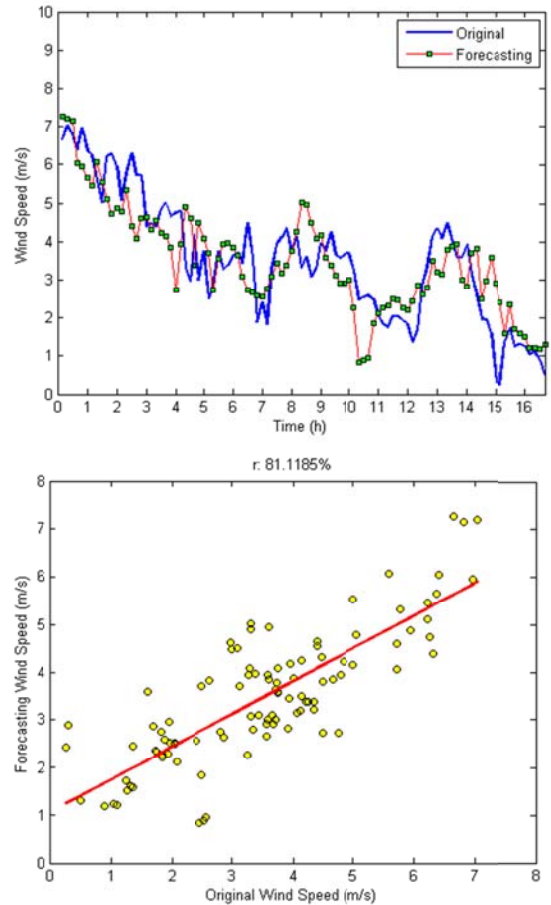

Fig. 9. 16 hours forward forecast and correlation of original data with forecast data, respectively.

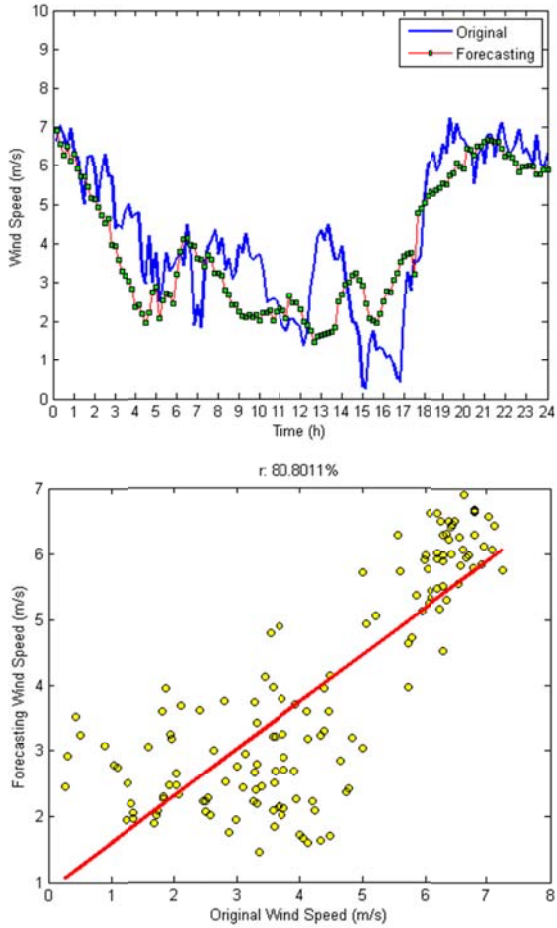

Fig. 10. 24 hours forward forecast and correlation of original data with forecast data, respectively.

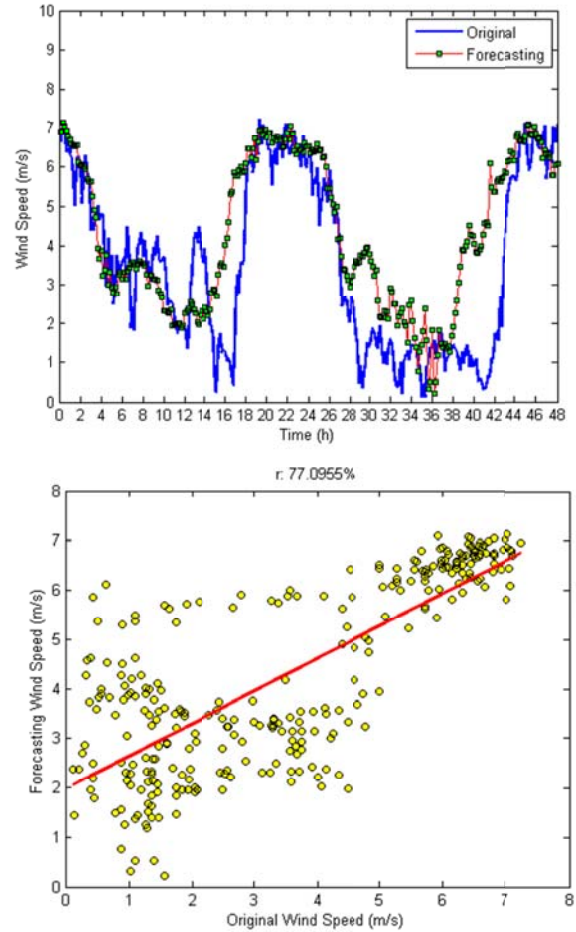

Fig. 11. 48 hours forward forecast and correlation of original data with forecast data, respectively.

The computational times for training and for prediction should be reduced, which enables examining more layers or neurons. Future work is the design and implementation of specialized hardware architectures on reconfigurable computing. Nowadays, several works in artificial intelligent techniques are embedded into FPGAs, for example, a predictor for global solar irradiation [17]. Moreover, different network schemes and levels of precision can be explored by using hardware architectures.

## 7. CONCLUSIONS

One of the most important steps in hybrid neuro-fuzzy modeling is the fuzzy membership values definition. As previously mentioned, the generalized Bell membership functions specified by four parameters were used in the present model.

There are two types of membership function: Gaussian and Bell membership functions. The Bell membership function has some advantages such as being a little more flexible than the Gaussian membership functions. Therefore the parameters of ANFIS would be better adjusted by using the Bell membership function. Also, both membership functions have advantages such as being smooth and non-zero at all points. In order to test the performance of ANFIS after training, the tested data were presented to the ANFIS. Each predicted value was compared against the actual observed value to measure the network performance. The coefficient of determination r gives information about the training of network, having a value of between [0, 100]. If the coefficient of determination is close to (100), it shows how successful the learning is. *MSE* is used to determine how much the network has reached the desired output values. In the particular case of the Isthmus of Tehuantepec the results from three models were presented. In each of them, data sets were constructed to carry out wind velocity forecasts.

Each model made forecasts within periods of 16, 24 and 48 forward hours. This involved wind velocity, barometric pressure, temperature and date as variables.

With regard to the forecast periods, one can conclude that error increase as the forecast period increases. Training with the first model, which had 17,320 data samples, for a 16 hours forward forecast produced $D=1$, $\Delta=1$ y $P=100$ since $100*10=1000$, and $1000/60=16$ hours, producing $r=81.1185$.

Subsequently, training with a second model with 17,273 data samples was performed for a 24 hours forward forecast period, producing $D=1$, $\Delta=1$ and $P=144$, since $144*10=1440$, and $1440/60=24$ hours producing $r=80.8011$ with a difference of 0.3174.

Later training was performed with a model containing 17,132 data samples for a 48 hours forward forecast that produced $D=1$, $\Delta=1$ y $P=288$, since $288*10=2880$, and $2880/60=48$ hours producing $r=77.0955$ and an incremental difference of 4.023.

It is worth mentioning that the three models were trained within six epochs. The training time was less for a hybrid like ANFIS than for a neuronal network. It is necessary to add that for future work, the utilization of other variables such as solar radiation, humidity, wind direction, etc., has not been discarded. The better performance of ANFIS with regard to the other intelligent methods is due to the combination of FL and ANN. Both mentioned membership functions (Bell and Gaussian) have been tested. It is important to mention that the rules used are generally based on the model and variables which depend on user's experience and trial and error methods.

Furthermore, the shape of membership functions depends on parameters, and changing these parameters will change the shape of the membership function. These problems are also seen in neural networks. Most of their parameters in a neural network are selected by trial and error method and most of them are dependent on the user's experience.

## ACKNOWLEDGMENT

The authors would like to thank to the PROMEP project 103.5/11/5266, UNISTMO-PTC-056 for providing financial support.

**Ernesto Cortés Pérez** received his M.S. degree in Computer Science, from ITA-LITI (Laboratory research on intelligent technologies) in Apizaco, Tlaxcala, Mexico. Since 2007 he has been Professor- Research at the University Isthmus, Oaxaca, Mexico. His current research interests include Intelligent Systems, Fuzzy Logic, Patterns Classification, Artificial Neuro-Fuzzy Networks, Bio-Inspired Algorithms and Artificial Vision.

**lgnacio Algredo-Badillo** received the B.Eng in Electronic Engineering from Technologic Institute of Puebla (ITP) in 2002 and the M.Sc and Ph.D degrees in Computer Science from National Institute for Astrophysics, Optics and Electronics (INAOE) in 2004 and 2008, respectively. Since 2009, he has been professor of Computer Engineering at University of Istmo. He has involved in the design and development of digital systems, reconfigurable architectures, software radio platforms, cryptographic systems, FPGAs implementations, microcontrollers-based systems and hardware acceleration for specific applications.

**Victor Hugo Garcia Rodriguez** received the M. Sc. degree in Electronic Engineering from Universidad de las Americas Puebla (UDLAP) in Cholula, Puebla, Mexico. Since 2002 he has been research professor at the University of Isthmus, Oaxaca, Mexico. His current research interests include power electronics, instrumentation and control.